%% file: elsarticle-template-harv.tex
\documentclass[preprint,12pt,authoryear,nonatbib]{elsarticle}

\usepackage{amssymb}
\usepackage{amsmath}

\usepackage[style=numeric,sortcites,backend=bibtex]{biblatex}
\usepackage{color,soul}
\usepackage{float}
\usepackage{multirow}
\usepackage{adjustbox}
\usepackage{subcaption}
\usepackage{graphicx}

\journal{Nuclear Physics B}

\makeatletter
\renewcommand{\@elsarticlehighlightstitle}{}
\makeatother

\makeatletter
\def\elsarticleprelims{}
\makeatother

\begin{document}

\begin{frontmatter}

\title{Hierarchical Reinforcement Learning in Multi-Goal Spatial Navigation with Autonomous Mobile Robots} 


\author{Brendon Johnson\corref{cor1}}
\author{Alfredo Weitzenfeld} 

\cortext[cor1]{Corresponding author. Email: brendon45@usf.edu, Phone: 5054278329}

\affiliation{organization={University of South Florida},
            addressline={4202 E. Fowler Avenue}, 
            city={Tampa},
            postcode={33620}, 
            state={Florida},
            country={United States of America}}

\begin{abstract}

Hierarchical reinforcement learning (HRL) is hypothesized to be able to leverage the inherent hierarchy in learning tasks where traditional reinforcement learning (RL) often fails. In this research, HRL is evaluated and contrasted with traditional RL in complex robotic navigation tasks. We evaluate unique characteristics of HRL, including its ability to create sub-goals and the termination functions. We constructed a number of experiments to test: 1) the differences between RL proximal policy optimization (PPO) and HRL, 2) different ways of creating sub-goals in HRL, 3) manual vs automatic sub-goal creation in HRL, and 4) the effects of the frequency of termination on performance in HRL. These experiments highlight the advantages of HRL over RL and how it achieves these advantages.  


\end{abstract}


\begin{highlights}
\item A comparison between standard reinforcement learning and hierarchical reinforcement learning.
\item An evaluation of different ways to create sub-goals.
\item A comparison between manual and automatic sub-goal creation.
\item A study of the frequency of sub-goals on performance.
\end{highlights}


\end{frontmatter}




\input{introduction}

\input{litreview}

\input{model}

\input{experiment_setup}

\input{experiments}

\input{con_discussion}

\printbibliography

 \end{document}

%% file: introduction.tex
\section{Introduction}
\label{intro}
Navigation is an essential part of robotics. Navigation problems can be extremely complex. This research aims to find out how to deal with this complexity. The proposed solution to the navigation problem involves reinforcement learning (RL), but it can struggle with tasks that have sparse reward structures \cite{dqn_human} \cite{hindsight_rl} \cite{reverse_rl} \cite{hindsight_replay}. 
Traditional RL is unreliable and often fails when solving tasks with long-term reward schemes. If the rewards in a task are few with many actions in between, as in a long-term reward scheme, RL will often be unreliable when temporal difference is used to train the algorithm. Temporal difference can fail in these tasks as the reward will decay, causing the temporal difference equations to fail because the reward signal becomes too small. Additionally, when using neural networks as part of the algorithm the weakened reward signal can cause decaying gradients, preventing sufficient learning. Another problem for RL in sparse reward tasks is that the algorithm needs to complete a complex sequence of actions before reaching a non-zero reward. This causes the agent to rarely receive positive rewards, as random exploration will rarely complete the sequence correctly \cite{hindsight_rl}.

One method to address this complexity is to build a hierarchy. This is inspired by nature as humans build hierarchies to solve problems. For example, if baking a cake, a person would first break it into parts: gathering the ingredients, mixing the ingredients, etc. This is a form of building a hierarchy by breaking down the task into smaller sub-tasks. This can be used in conjunction with RL to better solve complex tasks. 

Hierarchical reinforcement learning (HRL) is a form of reinforcement learning that aims to overcome some of the shortcomings of traditional RL in environments with sparse reward schemes by breaking down tasks into sub-tasks and creating a hierarchy. 
Creating this hierarchy can lead to several advantages over traditional RL, including faster training, improved adaptability, and the ability to solve sparse reward problems.

In this research, we investigate HRL’s ability to solve complex robot spatial navigation problems by creating sub-tasks while analyzing different sub-goal creation strategies. The paper provides the following analysis: 1) a comparison between traditional RL and HRL, 2) an evaluation of different methods for creating sub-goals, 3) a comparison of manual and automatic sub-goal creation, and 4) a study of the frequency of sub-goals on performance.

Each experiment is designed to highlight a different aspect of hierarchical reinforcement learning, thereby illustrating key components of HRL:
\begin{itemize}
    \item Experiment 1 is designed to show how HRL can solve sparse reward tasks when traditional RL fails. This is important because sparse reward tasks can be particularly difficult for RL, but many robotics tasks are sparse, so it is an important issue to study.
    
    \item Experiment 2 is designed to test different ways to create sub-goals and their importance in HRL. Although HRL was designed to build hierarchies to outperform standard reinforcement learning, some studies suggest this is not the reason for its improved performance \cite{HRLSometimes}. This experiment is designed to test if this assumption is correct. But to fully test this, more than one experiment was required. Experiment four is also designed to test this assumption.

    \item Experiment 3 is designed to evaluate manual versus automatic sub-goals creation. In most cases, manual sub-goals may be infeasible, but in certain cases, they may be practical and needed if they improve performance. The third experiment will assess whether they improve the performance of the agent.

    \item Experiment 4 is designed to test how the frequency of sub-goals impacts performance. This is important as it shows the optimal number of sub-goals. Additionally, in conjunction with experiment two, it demonstrates the importance of sub-goals in HRL algorithms. These experiments show this by evaluating the result of removing sub-goals from the algorithm. This is important as it explains why HRL is able to outperform traditional reinforcement learning in sparse reward tasks.
\end{itemize}

%% file: litreview.tex
\section{Literature Review}

\subsection{Hierarchical Reinforcement Learning}
There are many different types of reinforcement learning (RL) algorithms, including traditional “shallow” and hierarchical reinforcement learning (HRL). HRL is a type of reinforcement learning that breaks down tasks into sub-tasks, creating a hierarchy to solve a task. This design mimics how animals and humans break down tasks into smaller problems before completing a problem. In HRL, this design is applied to solving tasks with sparse reward structures, which are typically more challenging for traditional reinforcement learning algorithms \cite{dqn_human} \cite{hindsight_rl} \cite{reverse_rl} \cite{hindsight_replay}.

\subsection{Types of Hierarchical Reinforcement Learning}
There are multiple HRL algorithms, as summarized in Table \ref{tab:hrl_type} (see Shubham Pateria et al. \cite{HRLSurvey} for a review). 

\begin{table}[H]
    \centering
    \begin{adjustbox}{width=\columnwidth,center}
    \begin{tabular}{|c|c|c|}
        \hline
        \multicolumn{3}{|c|}{Hierarchical Reinforcement Learning} \\
        \hline
         & Sub-task Discovery & No Sub-task Discovery \\
        \hline
        \hline
        Single Agent, Single Task & Learning Hierarchical Policy and Sub-tasks & Learning a Hierarchical Policy \\
        & Sub-task Discovery & \\
        \hline
        Single Agent, Multiple Task & Transfer and Meta-learning for Sub-tasks & Transfer Learning without Sub-tasks  \\
        \hline
        Multiple Agent, Single Task & Multi-Agent Policies with Sub-task Discovery & Multi-Agent Policies \\
        \hline
    \end{tabular}
    \end{adjustbox}
    \caption{Different Types of Hierarchical Reinforcement Learning}
    \label{tab:hrl_type}
\end{table}

Our research is focused on single-agent, single-task problems with sub-task discovery. In this category, there are two main areas of research: the option architecture and feudal reinforcement learning \cite{SUTTON} \cite{FRL}. A third area is the MAXQ method \cite{MAXQ}, which has lost popularity over time as it has too many requirements. 

\subsection{Options}
\begin{figure}[H]
    \centering
    \includegraphics[width=0.5\linewidth]{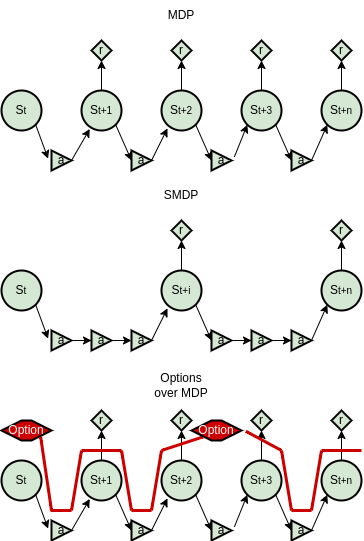}
    \caption{Difference Between MDP, SMDP, and MDP with Options where $s_t$ is the state at time $t$, a is the action at time $t$, and r is the reward at time $t$. At the top is a standard MDP, the middle is an SMDP where the actions take several time steps, and the bottom is how the option approximates an SMDP as an MDP.}
    \label{fig:option_diagram}
\end{figure}

One of the most popular types of Hierarchical Reinforcement Learning is the Option Architecture developed by Sutton et al. \cite{SUTTON}. 
The Option Architecture is a way to introduce temporal abstraction to a Markov Decision Process (MDP) \cite{sutton_rl_intro}. Introducing time violates the assumptions of an MDP, making such a process a Semi-Markov Decision Process (SMDP). SMDPs are difficult to solve. For this purpose, Sutton et al. \cite{SUTTON} introduced the Option Architecture, which allows for an SMDP to be approximated as an MDP while still incorporating temporal abstraction.
This is accomplished by replacing actions in the MDP with options, where an option is an action temporally extended over several states, as illustrated in Figure \ref{fig:option_diagram}. An option is comprised of three parts: the initiation set, the policy, and the termination set. The initiation set is a subset of the state space where the option can be selected and used. The policy is the actions taken during the selected option. The termination set is a subset of the state space where the option will end. An example of how this would look in robotics is if a robot vacuum is cleaning a house with three rooms, then each room would be an option, with the termination set being the door to the next room. The initiation set would be the door to the current room. The option policy would be the policy to clean the current room. Accordingly, each option is an independent MDP, with a policy over the options that is approximated as an MDP with the actions being options. The Option Architecture does not learn the sub-goals and assumes they are preset but there are many algorithms that build on top of the Option Architecture to solve this problem \cite{OptionCritic} \cite{TerminationCritic} \cite{DSC} \cite{automatic_discovery} \cite{qcut}.

\subsubsection{Option Critic Architecture}
The Option Critic Architecture builds upon the Option Architecture \cite{SUTTON} by incorporating an algorithm to create the options using equations based on the Policy Gradient Methods \cite{PolicyGradient}. It also uses an actor critic structure from the Actor-Critic Algorithms \cite{ACA} the options being the actor and critic used to evaluate them. To simplify the creation of the options, the architecture sets the initiation set to all states, so the initiation set does not need to be learned \cite{OptionCritic}. This has several advantages and disadvantages, the primary advantage being it simplifies the algorithm as it does not need to learn the initiation set, with the primary disadvantage being it removes some of the functionality of the Option Architecture, making the role of the policy over options more difficult as it has to evaluate every option in every state. There are several algorithms that are built on top of the Option Critic Architecture \cite{PPOOptionCritic} \cite{DSC}, including some discussed in future sections \cite{IOC} \cite{AOC}.

\subsubsection{Options of Interest}
The Interest Option Critic (IOC) Architecture is another algorithm built on top of the Option Critic Architecture \cite{OptionCritic}. When the Option Critic Architecture simplifies the problem by setting the initiation set to all possible states. The Interest Option Critic instead replaces the initiation set with a separate interest function for each option \cite{IOC}. This reduces the complexity but maintains the advantages of having an initiation set. Using the interest function makes the options specialized, resulting in faster learning and better adaptation \cite{IOC}.

\subsubsection{Attention Option Critic}
Attention Option Critic (AOC) is another algorithm that builds on top of the Option Critic Architecture \cite{OptionCritic} \cite{AOC}. It is similar to IOC \cite{IOC}, in that it adds some form of initiation set to the Option Critic Architecture \cite{OptionCritic}. In this case, instead of using the interest function, the AOC uses an attention mechanism to act as an initiation set and to determine when each option is used \cite{attenpaper}. This is done to make the options diverse, stable, and transferable \cite{AOC}.

\subsubsection{Deep Skill Chaining}
Although the Option Critic Architecture \cite{OptionCritic} is the predominant option architecture in the field of hierarchical reinforcement learning, another approach to using options is Deep Skill Chaining \cite{DSC}. Deep Skill Chaining is a form of HRL that chains options together to solve tasks and is based on Skill Chaining \cite{Skill}. Instead of initiation at the start state, DSC chains the skills backwards starting from the goal until it reaches the original state. In contrast to OC \cite{OptionCritic}, DSC does learn an initiation set for each option in the algorithm. This can have advantages such as making it so the option does not have to learn a policy for every state in the state space. In contrast, some disadvantages exist, such as making the algorithm more complex \cite{DSC}.

\subsection{Feudal Algorithms}
Another type of Hierarchical Reinforcement Learning is Feudal Reinforcement Learning \cite{FRL}. The goal of Feudal Reinforcement Learning is to speed up learning by learning at multiple resolutions in space and time \cite{FRL}. It does this by having multiple levels and learning at each one. It starts by creating a hierarchy with a high level manager who manages lower level sub-managers who, in turn, manage their own lower level managers. This goes on for as many layers as decided by the designer. The lowest level manager operates on a portion of the state space and outputs to action to the environment. The manager on top of the lowest level operates on an abstract state space and as levels increase in the state becomes more abstract. The higher level managers pick the lower level managers based on the their state abstractions. This structure can speed up learning and lead to better performance model.

\subsubsection{Feudal Networks for Hierarchical Reinforcement Learning}
The FeUdal Networks for Hierarchical Reinforcement Learning (FUN) is an algorithm built on top of the Feudal Algorithms \cite{FRL} \cite{FUN}. In this algorithm, instead of having a variable number of layers as in Feudal Reinforcement Learning, the FUN algorithm only has two levels. This is done to simplify the algorithm and make it have the same number of layers for any problem. In this algorithm, the top layer is known as the manager, and the lower level is known as the worker \cite{FUN}. The worker chooses the primitive actions in the environment and accesses the environment's state space directly. The manager works in an abstract state space and sends goals to the worker. The worker then uses the primitive actions to achieve the goal given by the manager. This allows the worker to work with short term goals, while the manager can focus on long term goals. This structure makes FUN better on long-term credit assignments than traditional algorithms.

\subsubsection{Hierarchical Reinforcement Learning With Off-Policy Correction}
HIerarchical Reinforcement learning with Off-policy correction (HIRO) is another algorithm with two levels of controllers \cite{HIRO}. The goal of HIRO is to be a general off-policy hierarchical reinforcement learning algorithm. This algorithm is similar to FUN \cite{FUN}, but one of the major differences is that HIRO does not abstract the state when communicating between the two levels \cite{HIRO}. The other major difference is that HIRO is an off-policy algorithm while FUN is on-policy.

\subsection{MAXQ}
The MAXQ Decomposition algorithm was a popular algorithm around the time it came out in the 1990s \cite{MAXQ}. The MAXQ algorithm requires the designer to define sub-goals and sub-tasks to complete these goals. The MAXQ then makes the hierarchy and uses it to decompose the value function. The value function is then used to create the policy for the task. 

%% file: model.tex
\section{HRL Architecture}
In this section, we describe the HRL architecture used for this research. We describe the policy gradient methods, options, option critic architecture, option critic hyperparameters, option degeneration, and the Proximal Policy Optimization (PPO) algorithm.

\subsection{Policy Gradient Methods}
Policy Gradient Methods \cite{PolicyGradient} are a popular family of reinforcement learning algorithms that derive equations to represent the policy with a function approximator. This algorithm is one of the first reinforcement learning algorithms and is the foundation for many state-of-the-art algorithms \cite{a3c} \cite{ddpg} \cite{pgq} \cite{mpo} \cite{sac}. The update function derived for this algorithm is shown in equation \ref{eq:gradient_policy} and is the basis of both the Option Critic Architecture \cite{OptionCritic} and Proximal Policy Optimization \cite{PPO}.

\begin{equation}
    \frac{\partial\rho}{\partial \omega} = \sum_s d^\pi(s)\sum_a \frac{\partial\pi(s,a)}{\partial\theta}
    f_\omega(s,a)
    \label{eq:gradient_policy}
\end{equation}
where
\begin{itemize}
    \item $s$ is the state
    \item $a$ is the action
    \item $f$ is the critic with parameters $\omega$
    \item $d^\pi(s)$ is the discounted weight of the states encountered
    \item $\pi$ is the policy with parameters $\theta$
    \item $\rho$ is the performance measure
\end{itemize}

\subsection{Options}
The options model used for this research is based on Options from Sutton, Precup, and Singh \cite{SUTTON}. In their work the authors proposed using options in reinforcement learning to build a hierarchy and introduce temporal abstraction into the algorithm. An option is based on the idea of an action over time; i.e., a temporally extended action. A policy can then use these options as actions and operate on a higher temporal plane while still acting like an MDP. 
Each option $\omega$, is a set of $(I_\omega,\pi_\omega,\beta_\omega)$. The $I_\omega$ is the initial set that determines when the option starts. The $\pi_\omega$ is the policy for the option, and the $\beta_\omega$ is the termination set determining when the option should end.

\subsection{Option Critic Architecture}
\begin{figure}[H]
    \centering
    \includegraphics[width=0.8\linewidth]{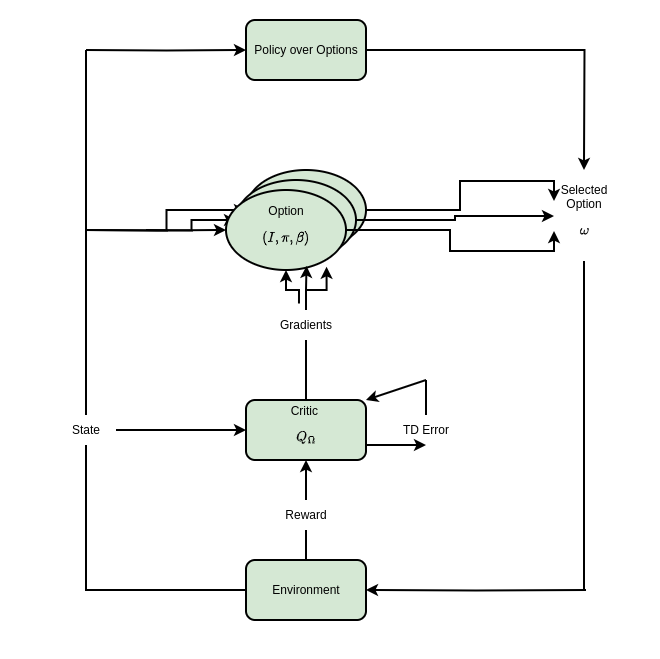}
    \caption{The Option Critic Architecture: Showing the Critic, the Options, and the Policy over Options. The Critic is used to evaluate the Options and create the gradients for them. The options vary in number as set by the hyperparameters. The Policy over Options chooses the option used in the current state.}
    \label{fig:oc_arch}
\end{figure}

The model used in these experiments is the Option Critic Architecture \cite{OptionCritic}. This architecture is built around options from Sutton et al. \cite{SUTTON}. The Option Critic Architecture has a select number of options, a policy over options, and a critic $Q_U$ as shown in Figure \ref{fig:oc_arch}. 

The critic evaluates an option, state, and action tuple and is the only part to be directly influenced by the reward. The loss function for the critic is constructed by finding the difference between the ground truth, $g_t$, and the current critic value. The ground truth is the sum of the current and future reward calculated using a modified version of the Bellman Equation to account for the options shown below in Equation \ref{eq:critic_update}.

\begin{equation}
    \begin{split}
        g_t=r_{t+1}+\gamma((1-\beta_{\omega_t,\upsilon}(s_t+1))\sum_a \pi_{\omega_t,\theta}(a\mid s_{t+1})Q_U(s_{t+1},\omega_t,a) +  \\
        \beta_{\omega_t,\upsilon}\max_\omega\sum_a \pi_{\omega,\theta}(a\mid s_{t+1})Q_U(s_{t+1}, \omega, a))
    \label{eq:critic_update}
    \end{split}
\end{equation}
where
\begin{itemize}
    \item $t$ is the current time step
    \item $\gamma$ is the rate of decay for the future rewards.
    \item $r_t$ is the reward at time step $t$
    \item $\beta_{\omega_t,\upsilon}$ is the termination function of option $\omega$ at time $t$, with the parameters $\upsilon$
    \item $\pi_{\omega,\theta}$ is the policy for option $\omega$ with parameters $\theta$
    \item $Q_U$ is the critic
\end{itemize}

Both the policy over options and the options make use of the critic. The policy over options uses the critic to decide which option is best in each state. To do this, it sums each value of each action multiplied by the likelihood of the action for the options, which is shown in Equation \ref{eq:q_omega}.

\begin{equation}
    Q_\Omega(s,\omega)=\sum_a\pi_{\omega,\theta}(a|s)Q_U(s,\omega,a)
    \label{eq:q_omega}
\end{equation}
where
\begin{itemize}
    \item $\omega$ is the current option
    \item $Q_U$ is the critic
    \item $Q_\omega$ is the total value of an option in state $s$
    \item $\pi_{\omega,\theta}$ is the policy for option $\omega$ with parameters $\theta$
\end{itemize}

The policy over options is a greedy algorithm that uses the critic to select the best option by calculating $Q_\Omega$ for each option and choosing the highest value. Exploration is introduced into this policy by using a decaying $\epsilon$ to randomly select an option instead of choosing the best. The $\epsilon$ will start at one and decay to zero over n steps.

The last part of the model is the options themselves. As stated before an option, $\omega$, consists of three parts the initiation set $I_\omega$, the policy $\theta_\omega$, and the termination function $\beta_\omega$. 
The initiation set, $I_\omega$, determines when the option starts. For this architecture, it is set to every state in the state space, eliminating the need to learn it. This means all options are available everywhere.
The option policy, $\theta_\omega$, is the policy learned for each option. These are the policies used to reach the sub-goal for each option. They effectively solve one of the sub-tasks for the algorithm. To learn these policies, a standard deep neural network is used with an update equation based on the policy gradient methods modified for options, as described in Equation \ref{eq:policy_update}. It uses the current critic value to evaluate the current policy and update it according to the critic.

\begin{equation}
    \theta \leftarrow \theta + \alpha_\theta \frac{\partial \log \pi_{\omega,\theta}(a\mid s)}{\partial\theta} Q_U(s,\omega, a) \label{eq:policy_update}
\end{equation}

\begin{figure}[H]
    \centering
    \includegraphics[width=0.8\linewidth]{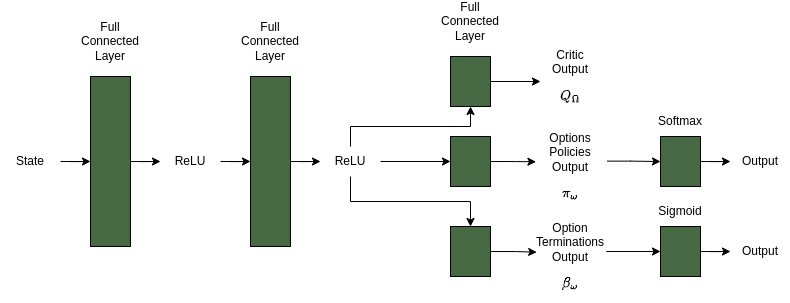}
    \caption{The Neural Network used for the Option Critic Architecture, with two shared layers feeding into separate heads for the critic, policy, and termination functions.}
    \label{fig:oc_nn}
\end{figure}

The termination function, $\beta_\omega$, determines when each option has reached its sub-goal and is terminated. This is important because it controls how long each option lasts and when to change options. A deep neural network is also used to approximate the termination functions. The update function uses the probability of termination multiplied by the value of the current option versus the other options, as described in Equation \ref{eq:term_update}.

\begin{equation}
    \upsilon \leftarrow \upsilon - \alpha_\upsilon \frac{\partial\beta_{\omega,\upsilon}(s')}{\partial\upsilon}(Q_\Omega(s',\omega) - V_\Omega(s')) + \phi
    \label{eq:term_update}
\end{equation}
where $V_\Omega$ is the option with the highest value, and $\phi$ is the termination regulation value. The termination regulation value is important because without it the options tend to terminate frequently so it was added to prevent degeneration. The neural network used for this algorithm is shown in Figure \ref{fig:oc_nn}. It has two shared layers which then feed into separate heeds for the critic, policy, and termination functions.

The Option Critic Architecture itself provides no guarantee of convergence, but it is built off of the Option Architecture, which guarantees the convergence of the option policy if every action in every state is executed infinitely often \cite{SUTTON}. There is one main reason this guarantee does not carry over into the Option Critic Architecture, which is that the Option Architecture does not learn the sub-goals and assumes they are set beforehand. Since the Option Critic Architecture learns the sub-goals, the guarantee of convergence does not apply. Meaning if the sub-goals are not correct, the algorithm may not converge, but in practice, this is rare as the algorithm is fairly robust.

\subsection{Option Critic Hyperparameters}
The hyperparameters used for these experiments are taken directly from Bacon, Harb, and Precup \cite{OptionCritic}. These parameters are shown in Table \ref{tab:parameters}.

\begin{table}[H]
    \centering
    \begin{tabular}{|c|c|}
        \hline
         \multicolumn{2}{|c|}{Option Critic Hyper-Parameters}  \\
         \hline
         \hline
         Learning rate $\alpha$ & 0.0005 \\
         Gamma $\gamma$ & 0.99 \\
         Termination Regularization $\phi$ & 0.01 \\
         Entropy Regularization & 0.01 \\
         Number of Options & 4 \\
         Update Frequency & 4 \\
         Freeze Interval & 200 \\
         Horizon & 1000 \\
         Max History & 10000 \\ 
         Batch Size & 32 \\
         Epsilon Decay & 50000 \\
         Temperature & 1 \\
         \hline
    \end{tabular}
    \caption{Hyper-Parameters Used For the Option Critic Architecture}
    \label{tab:parameters}
\end{table}

where
\begin{itemize}
    \item The Learning rate, $\alpha$, is the rate at which the parameters are changed. OC uses three different learning rates: $\alpha_\upsilon$, $\alpha_\theta$, and $\alpha_\Omega$, but to simplify the algorithm, all three have been set to the same value, $\alpha$.
    \item Gamma, $\gamma$, is the amount the future rewards are discounted.
    \item Termination Regularization, $\phi$, is the regularization term added to the termination loss function to prevent the option from terminating too frequently.
    \item Entropy Regularization is the regularization used to increase the policy entropy.
    \item Number of Options is how many options the algorithm uses.
    \item Update Frequency is how often the critic parameters are updated.
    \item Freeze Interval is how often the second critic is updated as two critics were implemented to prevent critic overestimation from Hasselt \cite{DoubleQ}.
    \item Horizon is how long the algorithm is allowed to explore before the environment is reset.
    \item Max History is the size of the replay buffer used by the algorithm.
    \item Batch Size is the size of the update batches.
    \item Epsilon Decay is when the epsilon decays to its lowest value by.
    \item Temperature is the temperature parameter for the softmax distribution.
\end{itemize}

\subsection{Option Degeneration}
\begin{figure}[H]
    \centering
        \begin{subfigure}{0.4\textwidth}
            \includegraphics[width=\linewidth]{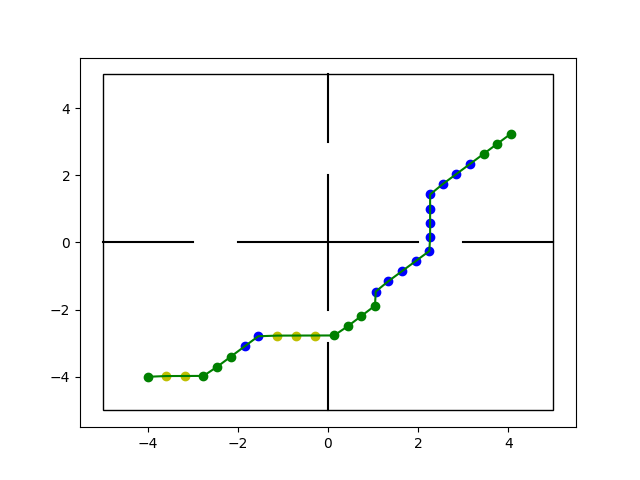}
            \caption{}
            \label{fig:path_norm}
        \end{subfigure}
        \hspace{1cm} 
        \begin{subfigure}{0.4\textwidth}
            \includegraphics[width=\linewidth]{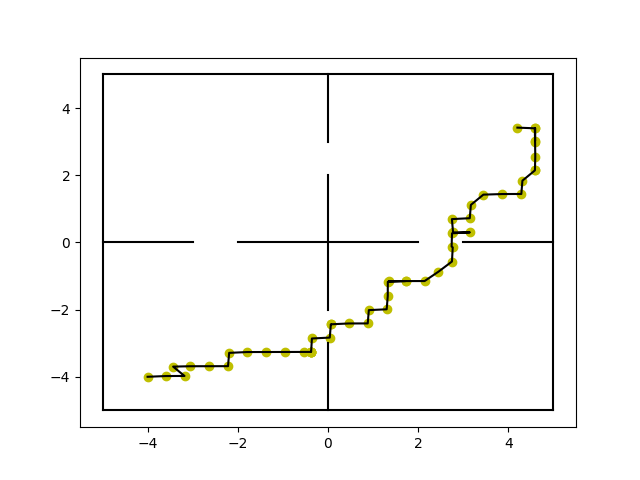}
            \caption{}
            \label{fig:path_one}
        \end{subfigure}
    \caption{Degenerated Option Example: a) non-degenerated sample path where each color is a different option, it is using three options in colors blue, green, and lime green, and b) degenerated option where every state is the same option, it is only using one option shown in lime green.}
    \label{fig:one_op}
\end{figure}

The primary disadvantage of the Option Critic Architecture is the options can degenerate if the hyperparameters are incorrect. There are two forms of degeneration in the Option Critic Architecture, which are the options shrinking or growing excessively. The first is when the options degenerate into a single action, this removes the advantages of the Option Critic because it makes the options essentially the same as the primitive action, making OC the same as standard RL. 
The second is when one option never terminates and grows to cover the entire task. This again makes the OC algorithm the same as standard RL because using only one option with one policy is the same as standard RL, removing the advantage of the OC. An example is shown in Figure \ref{fig:one_op}, where each dot is a state with the color corresponding to the option used. Figure \ref{fig:path_norm} shows a standard path where the options have not degenerated; as such, there are several different options being used. Figure \ref{fig:path_one} shows options that have degenerated and never terminate, with every state shown in the same color.

\subsection{Proximal Policy Optimization}
The Proximal Policy Optimization (PPO) algorithm \cite{PPO}, is a popular standard reinforcement learning algorithm. This algorithm is used as the baseline standard reinforcement learning algorithm to compare with hierarchical reinforcement learning. PPO is based on the policy gradient methods \cite{PolicyGradient} and trust region policy optimization \cite{trpo}, but is optimized to do multiple mini-batch updates instead of one update per sample. Another important part of PPO is the clipped surrogate loss in Equation \ref{eq:ppo_clip}, which prevents the algorithm from learning more than $1 + \epsilon$ in one step. To prevent it from taking too large of steps during convergence. This makes it an excellent algorithm in general and a perfect candidate to be a baseline as a state-of-the-art standard reinforcement learning algorithm \cite{ppo_a3c} \cite{ppo_trpo} \cite{ppo_car} \cite{ppo_carv2} \cite{ppo_uav}.

\begin{equation}
    L^{CLIP}(\theta)=E_t[min(r_t(\theta)A_t,clip(r_t(\theta), 1-\epsilon, 1+\epsilon)A_t)]
    \label{eq:ppo_clip}
\end{equation}
where
\begin{itemize}
    \item Epsilon, $\epsilon$, is a hyperparameter controlling the amount the loss is clipped.
    \item Theta, $\theta$, are the parameters for the policy.
    \item $A_t$ is the advantage at time $t$.
    \item $r_t$ is the reward at time $t$
\end{itemize}

\begin{figure}
    \centering
    \includegraphics[width=0.8\linewidth]{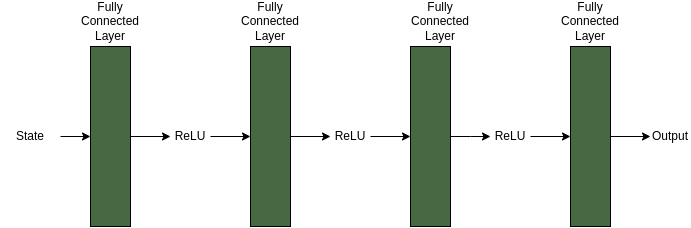}
    \caption{The Neural Network used for PPO with four fully connected layers and ReLU in between each.}
    \label{fig:ppo_nn}
\end{figure}

The neural network used for PPO in this research is shown in Figure \ref{fig:ppo_nn}. It has four fully connected layers with a ReLU in between each one \cite{relu_layer}.

%% file: experiment_setup.tex
\section{Experiment Setup}

\subsection{Task}
The task used in this research is the spatial navigation problem of navigating an unknown maze to reach a goal whose location is not known in advance. At each location, the robot has eight possible actions, each varying in its direction. A learning episode starts with the robot at the starting point set depending on the maze. At every step, the agent will choose an action and will receive a corresponding reward for its choice. This is repeated until the goal is reached or the number of steps has reached the set horizon to prevent an infinite amount of time in completing the learning task. The reward structure is critical in reinforcement learning, and hierarchical reinforcement learning is no exception. For this problem the reward structure is a large penalty for hitting a wall, a reward for reaching the goal, and a small penalty for any other state.

A multi-goal maze is defined as a maze with one primary goal and several sub-goals. The primary goal is set by the environment and gives a reward when reached. In comparison, the sub-goals are set by the algorithm, which has full control over them. This study tests the ability to create sub-goals of the Option Critic Architecture.

\subsection{Simulation Environment}
\begin{figure}
    \centering
    \includegraphics[width=0.8\linewidth]{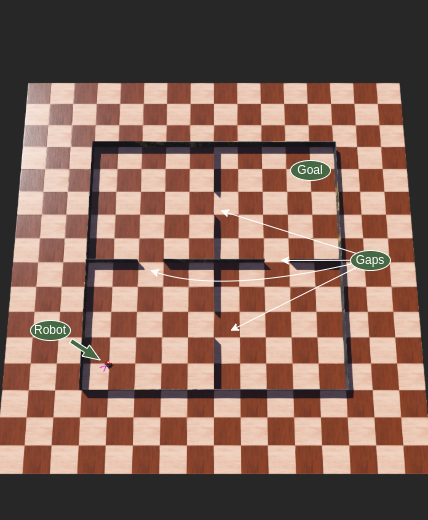}
    \caption{The four-rooms maze in the Webots simulator with the robot at the starting position. Each room has 2 gaps which serve as entry and exit points. Goal is at the opposite of the robot starting point.}
    \label{fig:webots}
\end{figure}

For the environment in these experiments, the Webots simulator was chosen along with the open-source FAIRIS framework to assist in robot control \cite{fairis}. An example of the Webots simulator is shown in Figure \ref{fig:webots}. This was chosen to simulate the real world as closely as possible when testing the algorithm. The University of South Florida's GAIVI supercomputer was used to perform the experiments, which on average took three days to complete. Additionally, all of the code and data for this paper are located in FAIRIS \cite{fairis}.

\subsection{Metrics}
Three metrics are used to evaluate the performance of the algorithm:
\begin{itemize}
    \item Path Length - number of actions the algorithms takes to reach the goal, and it measures how efficiently the algorithm can complete the task.
    \item Convergence Time - number of steps the algorithm takes to find the optimal path, i.e. when the algorithm stops learning and improving upon the optimal path.
    \item Average Option Length - number of steps before the option is terminated and a different option is chosen.
\end{itemize}

In the convergence time, each step is when the algorithm chooses an action and receives a state-reward pair from the environment. This is used instead of learning steps because some algorithms may be able to take multiple learning steps on a single set of data. Additionally, this equals the amount of real-world interaction the algorithm would need. The average option length metric is used to track both forms of option degeneration. The first form will show as the option length shrinks and the second as the option length grows.

\subsection{Exploration Structure}
For the Option Critic Architecture, the structure is to randomly sample from the distribution created by the current options policy. It is a decaying $\epsilon$ structure, so if a random number is less than $\epsilon$, then a random option is used; otherwise, the policy over options is used. The $\epsilon$ is then decayed over training time.

For the Proximal Policy Optimization \cite{PPO}, the exploration structure was set similar to the Option Critic Architecture. This structure was to randomly sample from the distribution created by the actor. As this was found not to explore enough and not to converge due to insufficient exploration, a version of the decaying $\epsilon$-greedy algorithm was used instead. This version was to sample a uniform distribution and if less than $\epsilon$ then a random action is chosen and $\epsilon$ is decayed. Otherwise, an action is chosen according to the PPO policy.

%% file: experiments.tex
\section{Experiments and Results}
Four different experiments were constructed to test the Option Critic Architecture: 1) a comparison of OC and PPO on three different mazes, each with increasing difficulty (one-room multiple small obstacles maze, one-room single large obstacle maze, four-rooms maze), 2) a comparison between critic created sub-goals and termination created sub-goals, 3) a comparison between manual and automatic sub-goal creation, and 4) an analysis of the frequency of terminations effect on performance. These experiments will give insight into hierarchical reinforcement learning compared to standard reinforcement learning and the impact of sub-goals on the algorithm.

\subsection{Mazes}
\begin{figure}[htbp]
    \centering
    \begin{minipage}{\textwidth}
        \centering
        \begin{subfigure}{0.4\textwidth}
            \includegraphics[width=\linewidth]{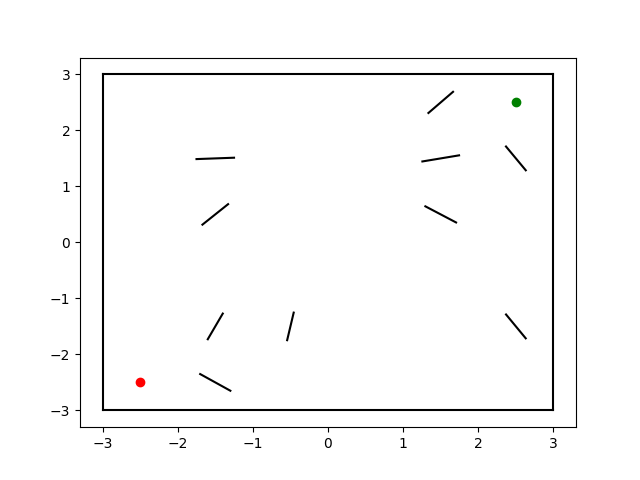}
            \caption{}
            \label{fig:maze_ten}
        \end{subfigure}
        
        \vspace{0.5cm}
        
        \begin{subfigure}{0.4\textwidth}
            \includegraphics[width=\linewidth]{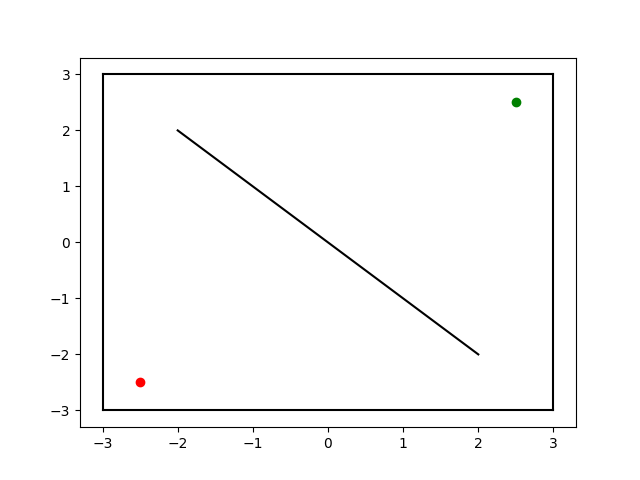}
            \caption{}
            \label{fig:maze_one}
        \end{subfigure}
        \hspace{1cm} 
        \begin{subfigure}{0.4\textwidth}
            \includegraphics[width=\linewidth]{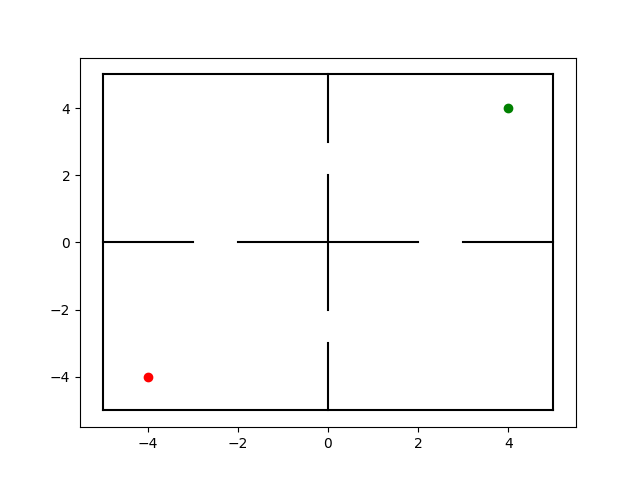}
            \caption{}
            \label{fig:maze_four}
        \end{subfigure}
    \end{minipage}
    \caption{All of the mazes used in the experiments: a) the one-room multiple small obstacles maze, b) the one-room single large obstacle maze, and c) the four-rooms maze (no obstacles). The red dot is the starting point for the robot in each maze and the green dot is the goal for each maze.}
    \label{fig:mazes}
\end{figure}

One of the standard benchmarks for hierarchical reinforcement learning is the four-rooms environment from Sutton et al. \cite{SUTTON}. In the experiments to compare OC with PPO, two open mazes with obstacles were used (one-room multiple small obstacles maze and one-room single large obstacle maze) and the four-rooms experiments, as shown in Figure \ref{fig:mazes}. The maze in Figure \ref{fig:maze_ten} is simple as there is only one room, and the obstacles are relatively small and randomly placed. The second maze in Figure \ref{fig:maze_one} is harder to solve, as it has one large obstacle that is placed directly between the agent and the goal, where the agent has to learn to go further away from the goal to be able to go around the obstacle and reach the goal. The third maze in Figure \ref{fig:maze_four} is the most difficult, corresponding to the four-rooms maze.

\subsection{Experiment 1: Comparison Between OC and PPO}
\begin{figure}[htbp]
    \centering
    \begin{minipage}{\textwidth}
        \centering
        \begin{subfigure}{0.4\textwidth}
            \includegraphics[width=\linewidth]{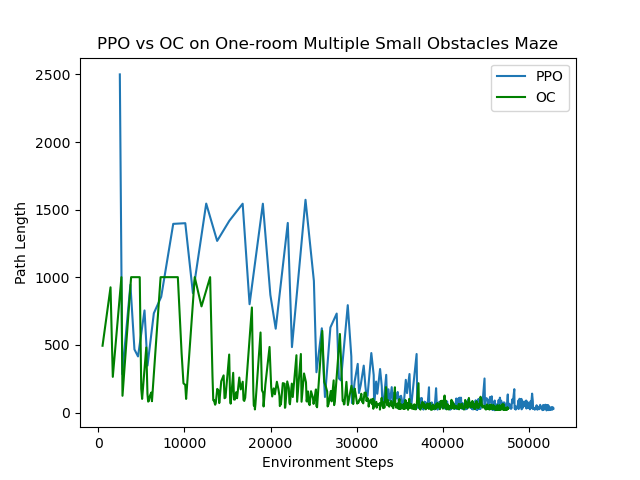}
            \caption{}
            \label{fig:ppo_ten}
        \end{subfigure}
        
        \vspace{0.5cm}
        
        \begin{subfigure}{0.4\textwidth}
            \includegraphics[width=\linewidth]{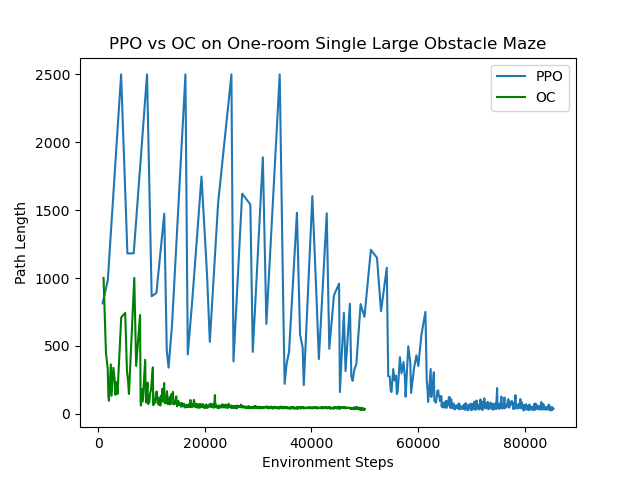}
            \caption{}
            \label{fig:ppo_one}
        \end{subfigure}
        \hspace{1cm} 
        \begin{subfigure}{0.4\textwidth}
            \includegraphics[width=\linewidth]{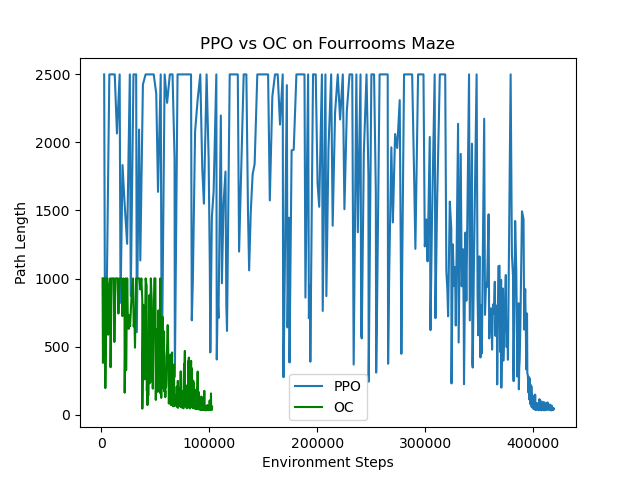}
            \caption{}
            \label{fig:ppo_four}
        \end{subfigure}
    \end{minipage}
    \caption{The results for the PPO vs OC for each maze: a) one-room multiple small obstacles maze, b) one-room single large obstacle maze, and c) four-rooms maze (no obstacles). The green line corresponds to OC, and the blue line is the PPO. As the mazes get more complex, the PPO performs worse when compared to the OC.}
    \label{fig:ppo_oc}
\end{figure}

This experiment is designed to test the difference in performance between the PPO and OC architectures. It is hypothesized that as the mazes get increasingly complicated, the PPO will become increasingly worse than the OC. Three different mazes are used with differing levels of difficulty to test the performance. They are a one-room multiple small obstacles maze, a one-room single large obstacle maze, and the four-rooms maze. 
In this experiment, the starting point is in the southwest corner, and the goal is in the northeast corner, as shown in Figure \ref{fig:mazes}, where the red dot is the starting point and the green dot is the goal. Each experiment is run until the algorithm converges or reaches a maximum number of steps to prevent an algorithm that won't converge from running forever. The maximum number of steps used in this experiment is 500000 steps. In each episode in the experiment, the robot is allowed to explore until it reaches the goal or the horizon. A learning step is completed after every simulation time step. Each experiment was run five times. A student's T-test was performed on the convergence time between the PPO and OC on each maze to see if the difference was statistically significant. The path length was too similar to warrant a statistical test.
The results of this experiment are shown in Figure \ref{fig:ppo_oc} and Table \ref{tab:ppo_oc_results}.

\begin{table}[H]
    \centering
    \begin{adjustbox}{width=\columnwidth,center}
    \begin{tabular}{|c|c|c|c|c|c|}
        \hline
        Maze & \multicolumn{3}{|c|}{Convergence Time} & \multicolumn{2}{|c|}{Path Length} \\
        \hline
        & PPO & OC & P-Value & PPO & OC \\
        \hline
        One-Room Multiple Small Obstacles Maze & 49191.0 & 44886.2 & 0.45 & 17.4 & 19.2 \\
        One-Room Single Large Obstacle Maze & 68684.4 & 42027.8 & 0.048 & 24.2 & 30.2 \\
        Four-Rooms Maze & 459396.8 & 175055.6 & 0.036 & 29.4 & 31.8 \\
        \hline
    \end{tabular}
    \end{adjustbox}
    \caption{PPO vs OC comparing path length and convergence time for each of the three mazes. Additionally, the results of a T-test between the PPO and OC convergence time for each maze are included to see if the results are significant.}
    \label{tab:ppo_oc_results}
\end{table}

In addition to converging faster, the OC model is about fifty times smaller than the PPO model shown in Table \ref{tab:model_params}.

\begin{table}[H]
    \centering
    \begin{tabular}{|c|c|c|c|}
        \hline
        Model & Number of Parameters & Layers & Layer Width \\
        \hline
        PPO & 1317897 & 4 & 256 and 512 \\
        OC & 25872 & 3 & 32 and 64 \\
        \hline
    \end{tabular}
    \caption{Number of Parameters for Both Models: Showing the difference in size between the models.}
    \label{tab:model_params}
\end{table}

Other model sizes were tested, but the final version was found to have the fastest convergence time. Each model size and convergence time are shown in Table \ref{tab:model_sizes}.

\begin{table}[H]
    \centering
    \begin{adjustbox}{width=\columnwidth,center}
    \begin{tabular}{|c|c|c|c|c|c|}
        \hline
        Model & Number of Layers & Layer Size & Total Parameters & Convergence Time & Path Length \\
        \hline
        OC & 3 & 32 and 64 & 25872 & 175055.6 & 31.8 \\
        PPO V1 & 3 & 128 and 256 & 330761 & 980848.0 & 29.0 \\
        PPO V2 & 3 & 256 and 512 & 792585 & 727093.0 & 31.0 \\
        PPO V3 & 4 & 128 and 256 & 462345 & 539333.0 & 29.0 \\
        PPO V4 & 4 & 256 and 512 & 1317897 & 459396.8 & 29.4 \\
        \hline
    \end{tabular}
    \end{adjustbox}
    \caption{All of the model sizes tested for the PPO with OC also shown to help compare.}
    \label{tab:model_sizes}
\end{table}

\subsection{Experiment 2: Difference Between Critic and Termination Automatically Created Sub-Goals}
\begin{figure}[htbp]
    \centering

    \begin{subfigure}{0.7\textwidth}
            \includegraphics[width=\linewidth]{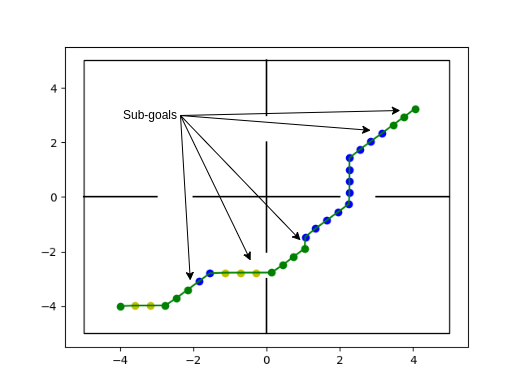}
            \caption{}
            \label{fig:standard_path}
    \end{subfigure}
    \vspace{1cm} 
    \begin{subfigure}{0.7\textwidth}
            \includegraphics[width=\linewidth]{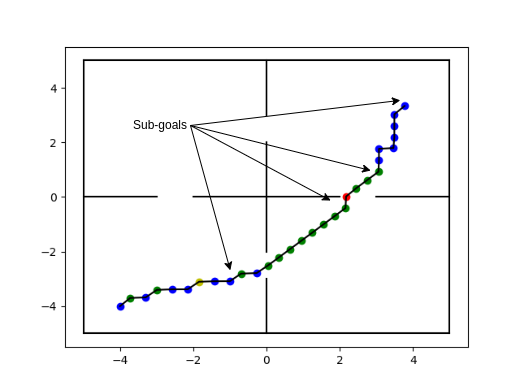}
            \caption{}
            \label{fig:no_term_path}
    \end{subfigure}
    
    \caption{The figure shows the path generated when termination created sub-goals are used compared to critic created sub-goals: a) termination created sub-goals path, and b) critic created sub-goals path. Every dot represents a state the robot reached on the path, and each color corresponds to the option used in that state.}
    \label{fig:no_term_comp}
\end{figure}

There are many different ways to automatically create the sub-goals. In this experiment, we compare critic created sub-goals and termination created sub-goals. Critic created sub-goals are where the critic repeatedly chooses the same option until reaching a certain sub-goal state. Although the policy over options chooses which option to use, the critic directly influences this decision, which is why they are called critic sub-goals. Termination sub-goals are the state where the termination function decides to end the current option. It is hypothesized that the termination created sub-goals will perform better than the critic created sub-goals because as shown in Equation \ref{eq:term_update} the termination function is updated to compare the differences between options to evaluate when to switch when in contrast as shown in Equation \ref{eq:critic_update} the critic is updated to evaluate an option in the current state.
The original OC model uses termination created sub-goals \cite{OptionCritic}. To compare to the critic created sub-goals, the termination function has been removed completely, and instead, the option will be terminated at every step. This is done to remove the termination function and prevent it from creating sub-goals, so the critic will set the sub-goals instead. This experiment will be tested on the four-rooms maze. This modified algorithm was then tested and compared to the original algorithm the results are shown in Table \ref{tab:criticvsterm} and Figure \ref{fig:no_term_comp}. A student's T-test was used on both the convergence time and path lengths to see if the differences were statistically significant, also shown in Table \ref{tab:criticvsterm}.

\begin{table}[H]
    \centering
    \begin{tabular}{|c|c|c|}
         \hline
         Sub-Goal Creation & Convergence Time & Path Length \\
         \hline
         Termination & 118489.85 & 31.5 \\
         Critic & 172863.95 & 31.8 \\
         P-Value & 3.35e-07 & 0.55 \\
         \hline
    \end{tabular}
    \caption{Experiment Four Results Showing the difference between the OC with termination created sub-goals and critic created sub-goals. Also shown is the result of a student's T-test between the convergence time and another test for the path lengths.}
    \label{tab:criticvsterm}
\end{table}

\subsection{Experiment 3: Comparison Between Automatic and Manual Sub-Goals}
This experiment is designed to test the differences between automatically set sub-goals and manually set sub-goals. In this experiment, the sub-goals will be created by the termination function and will be tested on the four-rooms maze.
It is hypothesized that the algorithm will perform better if it is able to create the sub-goals instead of them being set by hand. This is because the algorithm can set sub-goals to help explore at the beginning of the training. 

\begin{figure}[htbp]
    \centering
    \includegraphics[width=0.7\linewidth]{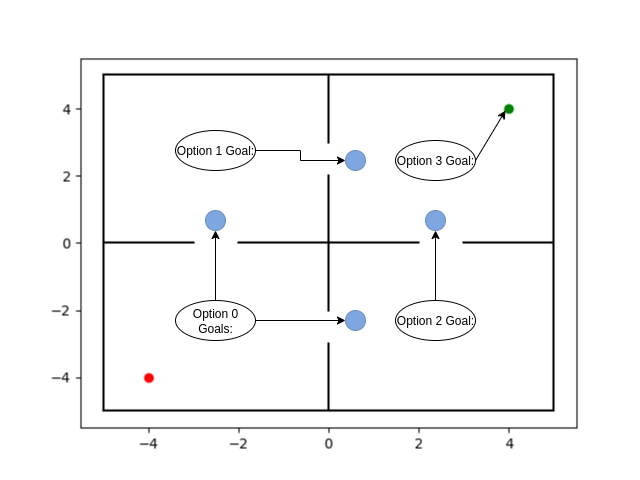}
    \caption{This figure shows the sub-goals for each option. The start point is shown in red, the goal in green, and the sub-goals in blue. Each option starts when the previous one ends.}
    \label{fig:handset}
\end{figure}

To test this hypothesis, the algorithm was modified to have the sub-goals set by hand. To do this, the termination code was replaced with code that checks if the algorithm is in a specific handset state using a cosine similarity and terminates the current option accordingly. Then, the states to terminate were with sub-goals that divide the task into equal parts to make each sub-task equal in length. For each option, the sub-goal was set to the doorway to the next room or the goal if the goal was in the room. Figure \ref{fig:handset} shows the sub-goals for each option. The doorways are used to break the problem down into four equal parts, dividing the work among the options. Although there are alternative ways to set the sub-goals, this was used because the four-rooms problem was designed to be easily broken down in this manner for testing hierarchical reinforcement learning algorithms \cite{SUTTON}.
Additionally, the option selection code was replaced to prevent the critic from choosing the wrong option for that room; instead, the option was selected according to the hand-set sub-goals. The doorways were chosen as the sub-goals because they even break up the problem as the four-room maze was designed to be broken down into rooms \cite{SUTTON}.

Each algorithm was run five times, and a student's T-test was performed on both the path length and convergence time shown in Table \ref{tab:exp_3_results}. This table shows that the model set sub-goals converge about twice as fast, and the paths are about ten steps shorter.

\begin{table}[H]
    \centering
    \begin{tabular}{|c|c|c|}
        \hline
        Sub-Goal Creation & Convergence Time & Path Length \\
        \hline
        Model Set & 163815.0 & 32.4 \\
        Hand Set & 350398.4 & 43.4 \\
        \hline
        P-Value & 0.00021 & 0.00141 \\
        \hline
    \end{tabular}
    \caption{Experiment Three: Table shows the difference in results between the model created sub-goals and hand set sub-goals.}
    \label{tab:exp_3_results}
\end{table}

\subsection{Experiment 4: Performance Evaluation on Frequency of Terminations}
\begin{figure} [H]
    \centering
    \includegraphics[width=1\linewidth]{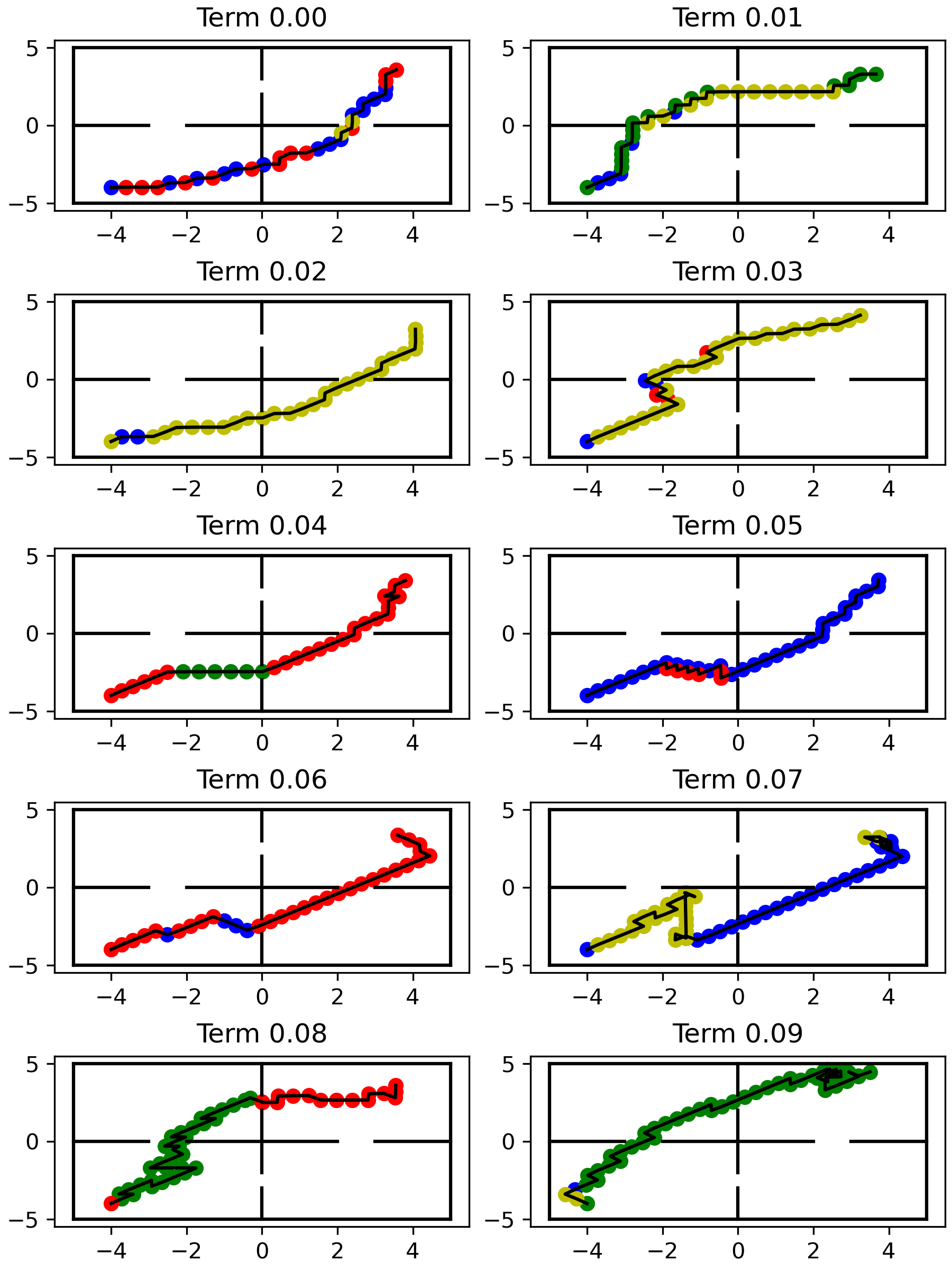}
    \caption{Termination Regulation Paths for each of the tested Termination Regulation values in Table \ref{tab:term_reg_data}, showing how the options terminate less frequently as the termination value increases.}
    \label{fig:reg_path}
\end{figure}

One of the expected strengths of applying Hierarchical Reinforcement Learning is its ability to break down tasks into sub-tasks and create sub-goals. In this experiment termination created sub-goals will be used. Sub-goals are an important part of the hierarchy created by the algorithm because they define how the task is divided into parts. Because of it's importance this experiment tests the effects of the frequency of termination on the performance of the algorithm. The frequency of termination determines how many sub-goals are created, as the sub-goal is the state an option reaches before switching to the next option. Additionally if the termination function does not create enough sub-goals then the problem won't be split up correctly and one option could end up covering the whole task, also known as degerenating.
The termination regulation, $\phi$ value was added to the termination function update equation to control the frequency of termination as shown in Equation \ref{eq:critic_update}.
It is hypothesized as the $\phi$ shrinks, the algorithm will be able to create better termination sub-goals because the smaller $\phi$ will prevent the options from growing too large and degenerating. 
To test this hypothesis, the Option Critic will be run on the four-rooms maze while varying the $\phi$. All the hyperparameters will be fixed while the $\phi$ is varied. $\phi$ will be varied between 0 and 0.1 with increments of 0.01.
These values were chosen to show how the performance changes as the options grow in size. In the original model, the value of 0.01 was used to incentivize creating larger options. Values larger than 0.01 were chosen to test whether incentivizing the creation of larger options is necessary.
This experiment will be performed on the four-rooms maze. The maximum number of steps used in this experiment is 500000 steps. This experiment was run five times to verify the results, and so a statistical test could be performed. An ANOVA test was used to test the differences between each termination value for each of the metrics: convergence time, option lengths, and path length. The results of this experiment are shown in Table \ref{tab:term_reg_data}. To better understand the results, several graphs of one of the experiments are constructed and shown below in Figures \ref{fig:op_lens}, \ref{fig:path_lens}, and \ref{fig:reg_path}. In Figure \ref{fig:op_lens}, the option lengths are shown for each of the termination regulation values. In Figure \ref{fig:path_lens}, the path lengths are shown as the algorithm is trained for each of the tested values. In Figure \ref{fig:reg_path}, the shortest path is shown for each of the test values.

\begin{table}[H]
    \centering
    \begin{adjustbox}{width=\columnwidth,center}
    \begin{tabular}{|c|c|c|c|}
         \hline
        Termination Regulation Value $\phi$ & Convergence Time & Average Option Lengths & Shortest Path  \\
         \hline
        0.00 & 91826.200 & 1.033 & 31.400 \\
        0.01 & 112609.800 & 4.960 & 32.000 \\
        0.02 & 145782.400 & 5.168 & 30.600 \\
        0.03 & 117765.000 & 6.814 & 31.600 \\
        0.04 & 189630.200 & 5.786 & 31.200 \\
        0.05 & 239664.800 & 8.655 & 33.800 \\
        0.06 & 422088.800 & 16.801 & 37.000 \\
        0.07 & 419348.400 & 24.593 & 42.600 \\
        0.08 & 500578.800 & 45.302 & 50.400 \\
        0.09 & 441668.200 & 36.145 & 46.800 \\
        0.10 & 500335.600 & 33.392 & 41.800 \\
        \hline
        P Value & 2.721e-10 & 7.927e-09 & 1.415e-06 \\
         \hline
    \end{tabular}
    \end{adjustbox}
    \caption{Termination regulation value $\phi$ results for convergence time, option lengths, and path length. The P-value is the result of the ANOVA test for each of the metrics in the table.}
    \label{tab:term_reg_data}
\end{table}

\begin{figure} [H]
    \centering
    \includegraphics[width=1\linewidth]{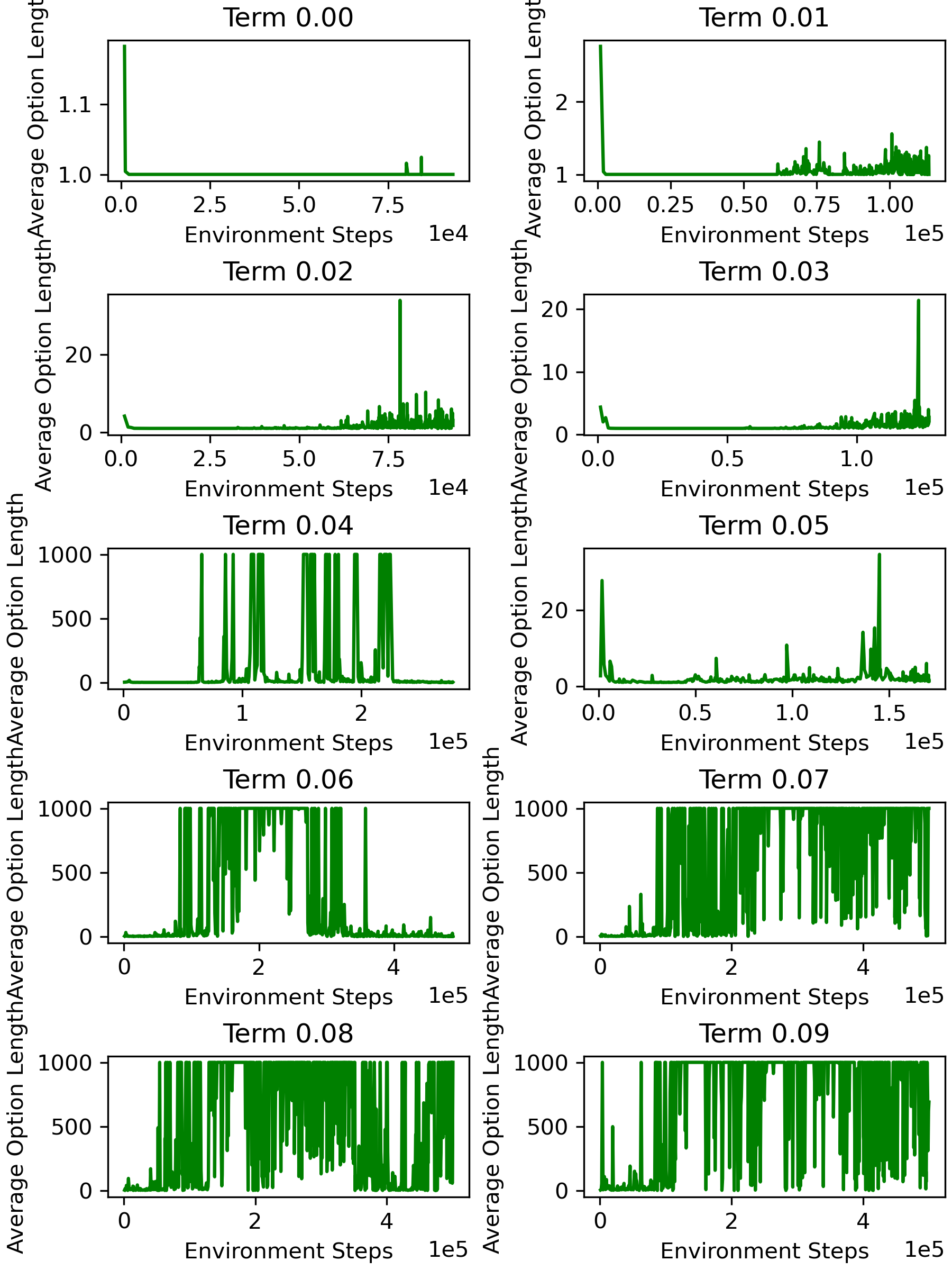}
    \caption{Termination Regulation Option Lengths during training for each of the tested Termination Regulation values in Table \ref{tab:term_reg_data}, showing how the option lengths increase as the termination value is increased.}
    \label{fig:op_lens}
\end{figure}

\begin{figure} [H]
    \centering
    \includegraphics[width=1\linewidth]{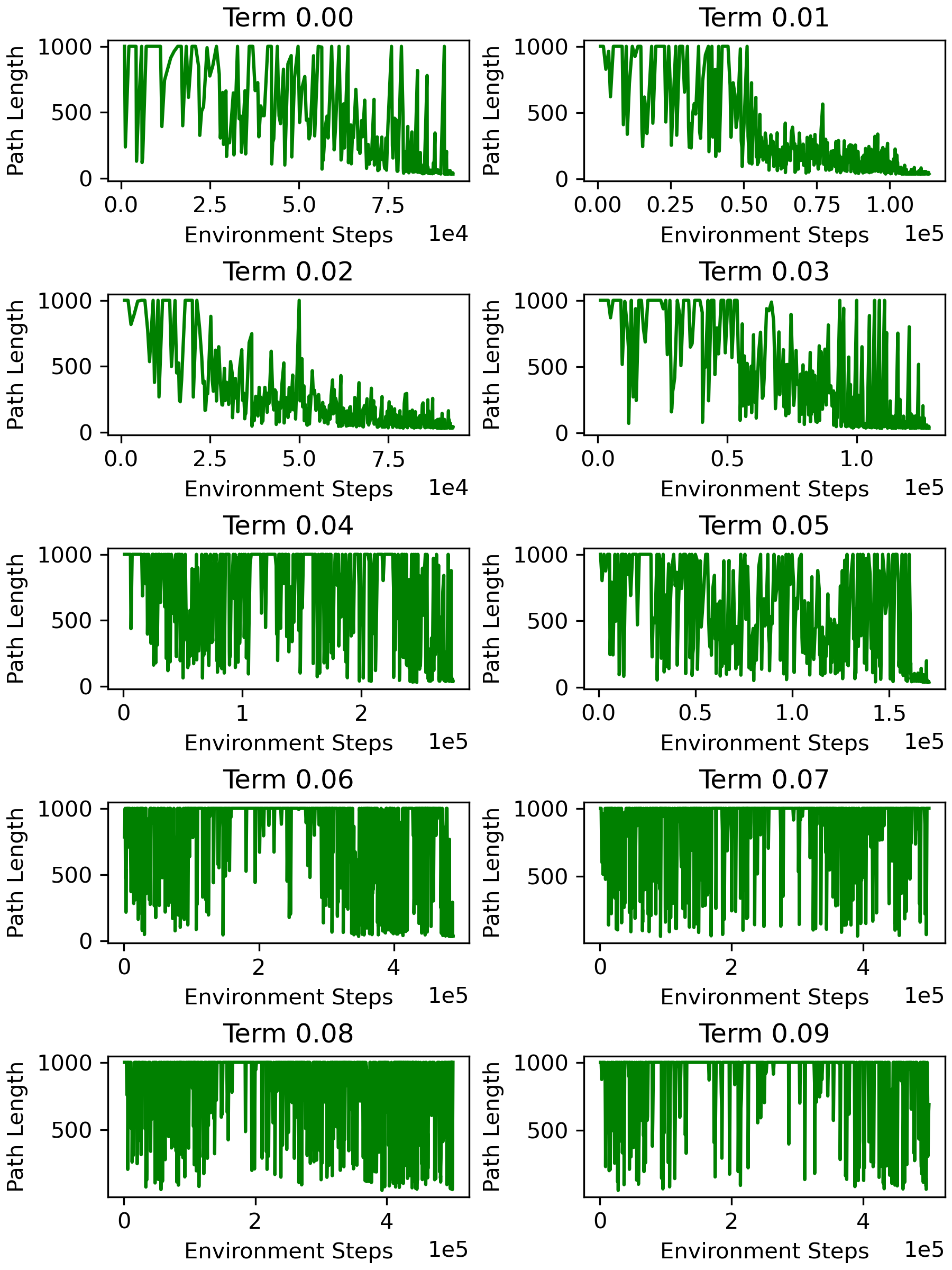}
    \caption{Termination Regulation Path Lengths during training for each of the tested Termination Regulation values in Table \ref{tab:term_reg_data}, showing how the time to converge increases as the termination value increases.}
    \label{fig:path_lens}
\end{figure}

%% file: con_discussion.tex
\section{Conclusions and Discussion}
Hierarchical Reinforcement Learning is a powerful tool that can solve complex problems that standard reinforcement learning struggles with. HRL is able to build a hierarchy that breaks down tasks into smaller sub-tasks by defining sub-goals. These sub-goals may have a significant impact on the performance of the algorithm and solve complicated problems. The primary contribution of this research is to demonstrate the importance of sub-goals. This research additionally has four secondary contributions to the scientific community: a comparison of standard reinforcement learning to hierarchical reinforcement learning, an evaluation of two different ways to automatically create sub-goals, a comparison of manual and automatically created sub-goals, and a performance evaluation of the frequency of termination.

In this section, we discuss the results of the experiments and show why each experiment is important. Experiment one highlights how HRL can outperform traditional reinforcement learning in sparse reward environments. Experiments two and four together demonstrate the importance of sub-goals in HRL, as well as investigating different methods and quantities of sub-goals. Experiment three is essential as it shows the importance of having the algorithm choose the sub-goals.

\subsection{Experiment 1: Comparison Between OC and PPO}
OC was shown to perform better when compared to PPO as the complexity increases. The results show that with one-room multiple small obstacles maze, the PPO has a similar convergence performance to the OC, and as the maze complexity increases, the convergence performance of PPO compared to OC decreases steadily. This is shown in the Table \ref{tab:ppo_oc_results} as the P-value for the one-room multiple small obstacles maze is 0.45, suggesting that the convergence performance is about the same. When the P-value for the one-room single large obstacle maze is less than 0.05 at 0.048, it is statistically significant, showing that the OC performs better than the PPO in terms of convergence time. With the four-rooms maze the difference is even larger where the OC converges about twice as fast as the PPO and has a P-value of 0.036. Although the convergence time is different between the PPO and OC, the path length is about the same. This was expected as they were both given time to converge, and they didn't have difficulties finding a near-optimal maze. Based on these results, a T-test was not performed on the path lengths.

In addition to being slower, PPO is fifty times larger than OC, primarily due to the PPO's network having an additional layer and the layers being eight times the size of the OC model shown in Table \ref{tab:model_params}. Three smaller PPO networks were tested shown in Table \ref{tab:model_sizes}. All four models had about the same final path lengths but had different convergence times. The fourth version of the PPO model or the final version was used in the comparison to Option Critic. The first version of the PPO has a fourth of the parameters of the final version but is significantly slower taking about double the time to converge. The second version has about half the parameters but is still much slower taking about 1.5 the number of steps as the final model to converge. The third version is only slightly slower than the final version taking about 1.15 the number of steps as the final model, but has a fourth of the parameters of the final model. Even though this version of the model is significantly smaller than the final PPO model it is still about ten times the size of the OC model.

Another conclusion we can draw from this experiment is OC's ability to outperform PPO as the rewards become increasingly sparse. In the one-room multiple small obstacle maze, the reward is relatively easy to achieve as the obstacles are small and can be avoided relatively easily, but as the rewards get harder to reach in the one-room single large obstacle and the four-rooms maze PPO's performance significantly decreases. This is because PPO can struggle as the rewards become increasingly sparse, but the OC is able to use the hierarchy it created to handle this sparsity.

\subsection{Experiment 2: Difference Between Critic and Termination Automatically Created Sub-Goals}
Our hypothesis was that the termination created sub-goals will perform better than critic created sub-goals. The results support this hypothesis as termination created sub-goals converged in about 50000 less steps than the critic created sub-goals as shown in Table \ref{tab:criticvsterm} and Figure \ref{fig:no_term_comp}. Although the convergence time is faster, the path lengths are about the same.
Although both models are able to set sub-goals, neither model is perfect, and this is shown in Figure \ref{fig:no_term_comp}. Both models sometimes set sub-goals too close together so that the option is changing every state or every couple of states. These are poor sub-goals because the algorithm is degenerating so that one option lasts a single state, which removes the purpose of the options. If the options last only a single state, then they are essentially the same as actions, removing the hierarchy and the purpose of using options.
These poor sub-goals are caused by three reasons: 1) the model is trained until it stops improving but this does not mean it has the best sub-goals and they might improve if it is trained for longer, 2) the algorithm explores by choosing a random option if a random number is greater than epsilon which decays over time but the algorithm generally converges before this decays to zero so the poor sub-goal may be the result of a random option being chosen, and 3) the learning algorithm is not perfect and may converge to a sub-optimal solution for example if it gets stuck in a local minima.
It is important to recognize that even though both models have imperfect sub-goals, the critic created sub-goals tend to be worse which is why it performs worse than the termination created sub-goals.

\subsection{Experiment 3: Comparison Between Automatic and Manual Sub-Goals}
The idea behind sub-goals is to break down tasks into smaller tasks, such that the smaller tasks are easier for the policies to solve as they are not as complicated. But it is important that the task is split evenly. If not, one of the tasks will be more difficult and would remove the purpose of creating the smaller tasks to make easier tasks. Since it is important to break down the tasks correctly, it may seem best to set the sub-goals by hand to ensure the task is broken down correctly. To manually set the best possible sub-goals for this experiment, each doorway and the goal were chosen as the sub-goals. These were chosen because they evenly break down the task. This is similar to the originally designed four-rooms maze task, where each room was assigned a different option, with the doorway being the sub-goals \cite{SUTTON}. As seen in Table \ref{tab:exp_3_results} when the algorithm uses the manually set sub-goals it takes about twice as long to converge and it's path length is about ten steps longer. These results support the hypothesis stating the algorithm will perform better if the sub-goals are set by the algorithm. These results support the hypothesis suggesting the algorithm is able to break down the tasks adequately, thus hand-set sub-goals are not required and even hinder the process as they impede exploration as most of the exploration in the Option Critic comes from trying different options and correspondingly pick a different action. The consideration is that if the sub-goals are set by hand, the algorithm wouldn't be considering different options in the same state, explaining why automatic sub-goals produce better results. In summary the automatic sub-goals perform better because they allow the algorithm to explore more not because the sub-goals are necessarily better. In this research there was no metric to compare the sub-goals to ideal sub-goals but for future research such metric could be used to better understand the differences between sub-goals created by algorithm and sub-goals set by hand.

\subsection{Experiment 4: Performance Evaluation on Frequency of Terminations}
The hypothesis was that as the $\phi$ shrinks, the algorithm will create better sub-goals because a smaller $\phi$ will create shorter options stopping them from growing too large. Meaning large values of $\phi$ such as 0.08-0.10 will fail to converge before the maximum cut off resulting in similar convergence times around 500000. The results support this hypothesis because as the $\phi$ increases, the convergence time increases as well. This is expected because as the options grow, they degenerate to cover the entire task, which eliminates the need for multiple options and removes the purpose of the Option Critic.
When analyzing the convergence time in Table \ref{tab:term_reg_data}, it shows that the termination regulation $\phi$ value of 0.00 has the shortest convergence time, corresponding to very short options. Additionally when $\phi$ is 0.00 the average option lengths are about one. This would suggest that the options terminate every state, but as shown in Figure \ref{fig:reg_path}, the options still can last longer than one state, resulting in the critic repeatedly picking the same option in those states. This suggests the critic assists the termination function in creating the sub-goals. This shows why the termination created sub-goals are better than critic created sub-goals because with termination created sub-goals the critic will still influence the sub-goals, and thus the critic and termination function are working together to create sub-goals.

Theoretically, sub-goals are advantageous because they enable an algorithm to break down tasks into several smaller, simpler sub-tasks, which are easier for the algorithm to solve. These results support this hypothesis because the Option Critic is able to outperform the PPO as shown in Table \ref{tab:ppo_oc_results}. But Option Critics' increased performance could come from other factors, in addition to sub-goals. 
In Table \ref{tab:term_reg_data}, it shows how as the termination regulation value increases from 0.00 to 0.10, the convergence time also increases. In Figure \ref{fig:path_lens}, it shows how as the termination regulation value increases, the sub-goals are longer and there are fewer sub-goals, up to the point where there are almost no sub-goals. In general, as the number of sub-goals decreases, the performance decreases, showing how it is important to have sub-goals. This might suggest that it would be best to terminate in every state, but as seen in experiment 3, this can also lead to poor performance.

\subsection{Impact of Sub-goals and the Hierarchy}
Current research into HRL has demonstrated its ability to outperform traditional reinforcement learning; however, there is a lack of research into why HRL performs better. Theoretically, it performs better due to its design to create a hierarchy for the task, but there is limited research on whether this is true in practice. Our research provides evidence of how sub-goals and building a hierarchy cause improvements in HRL.

Here, we demonstrate the importance of sub-goals by evaluating what happens when we remove them. There are two ways to remove the sub-goals: to remove the terminations and have one option or to terminate every step. The first option is essentially the same as traditional reinforcement learning, as a single option without terminations is a standard actor-critic policy. As discussed from the results of experiment one this results in worse performance. Additionally, in experiment four, we can see in Table \ref{tab:term_reg_data} that as the termination regulation value increases, the number of sub-goals decreases, and as a result, the performance decreases. 
This shows why it is important to have at least some sub-goals but additional analysis shows how as the termination regulation value decreases the number of sub-goals increases and the performance increases as well. This suggests that the performance improvements are not coming from the sub-goals, but rather from terminating and switching options or policies. To test this we created experiment two, in experiment two we terminate every step to see if terminating the options is causing the performance increase. From experiment two's results we can learn two things first the termination is not were the performance increase comes from, as terminating every step performs worse. The second is that even if we remove the termination functions and terminate every step instead; the critic will still create sub-goals. This explains why in experiment four the frequent terminations performed better because the best performance is when the termination function and the critic work together to create the sub-goals. Which is not possible if the options do not terminate frequently, as the critic is only used when the options are terminated. These experiments show the importance of sub-goals because if we remove them by not terminating or terminating every step then the performance decreases.

This illustrates the importance of sub-goals in most situations, but this may not be true in all cases. As there may be simple problems that do not need to be broken up and thus do not require sub-goals. Furthermore, in simple problems, sub-goals may hinder performance by adding unnecessary complexity to the task.
By showing the importance of sub-goals, we also show the importance of the hierarchy since creating sub-goals creates a hierarchy as the sub-goals define when each sub-task ends and the next begins.



\subsection{Future Work}
Future research will investigate further ways to set the sub-goals automatically. Currently, the policy over options relies mostly on the critic to create sub-goals, but this could be replaced by using a different algorithm for the policy over options, such as an additional network or a different machine learning technique. This could improve the exploration of the sub-goals and the overall algorithm performance. Future research will also investigate the performance of the Option Critic in physical robot learning tasks in spatial navigation tasks in the context of sequence learning \cite{cazin} \cite{medeiros} \cite{Scleidorovich2022-zd}.